\documentclass[11pt,a4paper]{article}
\usepackage{emnlp-ijcnlp-2019}
\usepackage{CJK,algorithm,algorithmic,amssymb,amsmath,array,epsfig,graphics,multirow,array,float,subfigure,verbatim,epstopdf}
\usepackage{times}
\usepackage{latexsym}
\usepackage{enumitem}
\usepackage{color}
\usepackage{hhline}
\usepackage{booktabs}
\usepackage{flushend}
\usepackage{amssymb}
\usepackage{xspace}
\usepackage{subfigure}
\usepackage{xcolor,colortbl}

\usepackage{url}

\usepackage{pifont}
\newcommand{\cmark}{\ding{51}}%
\newcommand{\xmark}{\ding{55}}%

\newcommand{\papername}{\textsc{Cosmos}\xspace}
\newcommand{\dataname}{\textsc{Cosmos} QA\xspace}

\aclfinalcopy 

\title{
\papername QA: Machine Reading Comprehension 
\\with Contextual Commonsense Reasoning 
}

\author{Lifu Huang$^{\spadesuit}$\thanks{\textsuperscript{*}The work has been done during the author's internship in AI2.}, Ronan Le Bras$^{\clubsuit}$, Chandra Bhagavatula$^{\clubsuit}$, Yejin Choi$^{\clubsuit,\diamondsuit}$ \\
  $^{\spadesuit}$ University of Illinois Urbana-Champaign, Champaign, IL, USA \\
  $^{\clubsuit}$ Allen Institute for Artificial Intelligence, Seattle, WA, USA \\
  $^{\diamondsuit}$ University of Washington, Seattle, WA, USA \\
  {\tt lifuh2@illinois.edu } \\
  {\tt \{ronanlb, chandrab, yejinc\}@allenai.org} 
 }

\date{}

\begin{document}
\maketitle

\begin{abstract}
Understanding narratives requires reading between the lines, which in turn, requires interpreting the likely causes and effects of events, even when they are not mentioned explicitly. In this paper, we introduce \dataname, a large-scale dataset of $35,600$ problems that require commonsense-based reading comprehension, formulated as multiple-choice questions. In stark contrast to most existing reading comprehension datasets where the questions focus on factual and literal understanding of the context paragraph, our dataset focuses on reading between the lines over a diverse collection of people's everyday narratives, asking such questions as \emph{``what might be the possible reason of ...?''}, or \emph{``what would have happened if ...''} that require reasoning beyond the exact text spans in the context. To establish baseline performances on \dataname, we experiment with several state-of-the-art neural architectures for reading comprehension, and also propose a new architecture that improves over the competitive baselines. Experimental results demonstrate a significant gap between machine (68.4\%) and human performance (94\%), pointing to avenues for future research on commonsense machine comprehension. Dataset, code and leaderboard is publicly available at \url{https://wilburone.github.io/cosmos}.
\end{abstract}

\section{Introduction}

\begin{figure}[t!]
\centering
\includegraphics[width=0.48\textwidth]{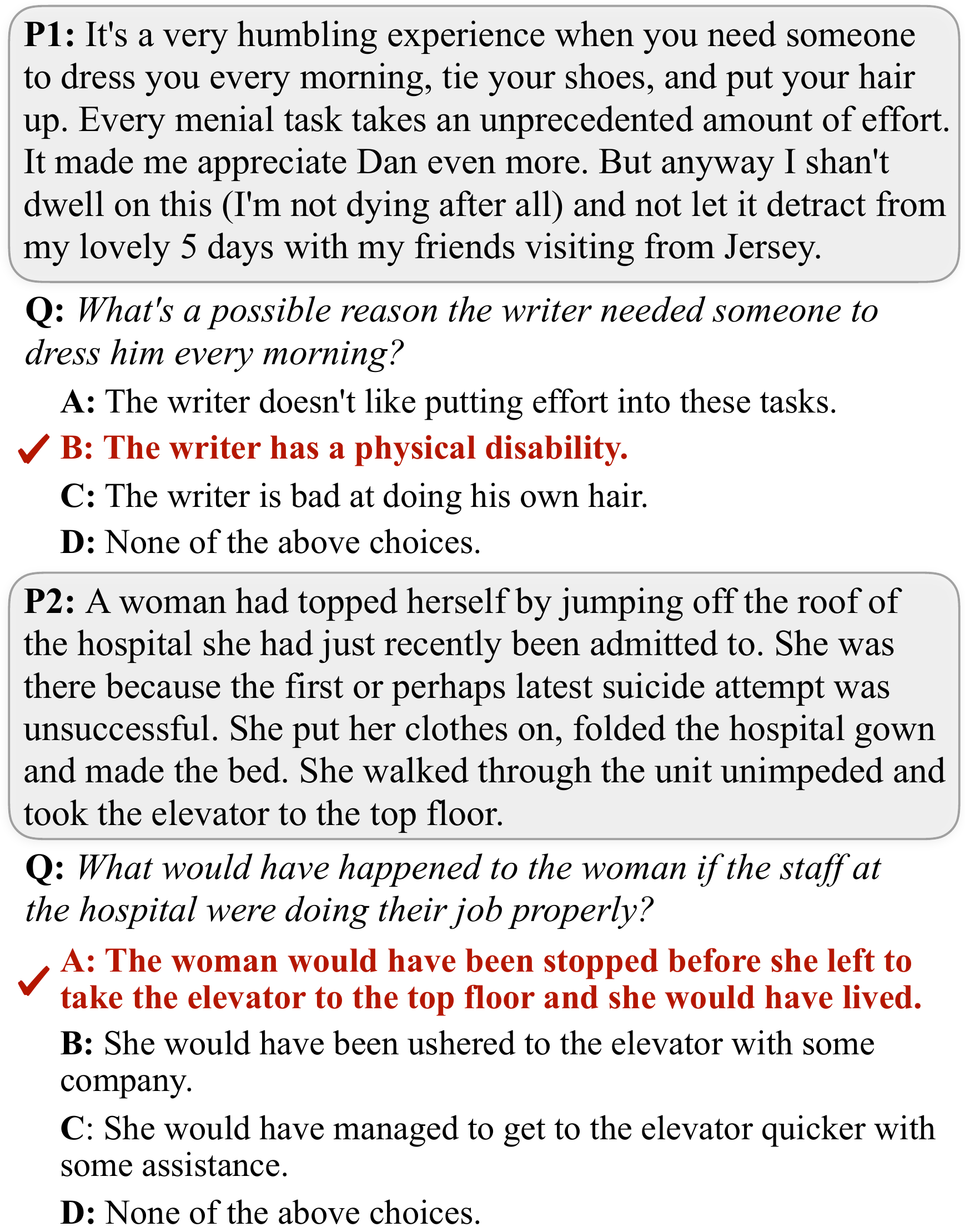}
\caption{Examples of \dataname. 
({\cmark} indicates the correct answer.) 
Importantly, (1) the correct answer is not explicitly mentioned anywhere in the context paragraph, thus requiring reading between the lines through commonsense inference and (2) answering the question correctly requires reading the context paragraph, thus requiring reading comprehension and \emph{contextual} commonsense reasoning.}
\label{fig:example}
\vspace{-0.5cm}
\end{figure}

Reading comprehension requires not only understanding what is stated explicitly in text, but also \emph{reading between the lines}, i.e., understanding what is not stated yet obviously true \cite{Norvig:CSD-87-339}. 

For example, after reading the first paragraph in Figure~\ref{fig:example}, we can understand that the writer is not a child, yet needs someone to dress him or her every morning, and appears  frustrated with the current situation. Combining these clues, we can infer that the plausible reason for the writer being dressed by other people is that he or she may have a physical disability. 

As another example, in the second paragraph of Figure~\ref{fig:example}, we can infer that the woman was admitted to a psychiatric hospital, although not mentioned explicitly in text, and also that the job of the hospital staff is to stop patients from committing suicide. Furthermore, what the staff should have done, in the specific situation described, was to stop the woman from taking the elevator.

There are two important characteristics of the problems presented in Figure~\ref{fig:example}. First, the correct answers are not explicitly mentioned anywhere in the context paragraphs, thus requiring reading between the lines through commonsense inference. Second, selecting the correct answer requires reading the context paragraphs. That is, if we were not provided with the context paragraph for the second problem, for example, the plausible correct answer could have been B or C instead.  

In this paper, we focus on reading comprehension that requires \emph{contextual} commonsense reasoning, as illustrated in the examples in Figure~\ref{fig:example}. Such reading comprehension is an important aspect of how people read and comprehend text, and yet, relatively less studied in the prior machine reading literature.
To support research toward commonsense reading comprehension, we introduce \papername QA (\underline{\textbf{Co}}mmon\underline{\textbf{s}}ense \underline{\textbf{M}}achine C\underline{\textbf{o}}mprehen\underline{\textbf{s}}ion), 
a new dataset with $35,588$ reading comprehension problems that require reasoning about the causes and effects of events, the likely facts about people and objects in the scene, and hypotheticals and counterfactuals. Our dataset covers a diverse range of everyday situations, with $21,886$ distinct contexts taken from blogs of personal narratives. 

The vast majority ($93.8\%$) of our dataset requires contextual commonsense reasoning, in contrast with existing machine comprehension (MRC) datasets such as SQuAD~\cite{rajpurkar2016squad}, RACE~\cite{lai2017race}, Narrative QA~\cite{kovcisky2018narrativeqa}, and MCScript~\cite{ostermann2018mcscript}, where only a relatively smaller portion of the questions (e.g., $27.4\%$ in MCScript) require commonsense inference. 
In addition, the correct answer cannot be found in the context paragraph as a text span, thus we formulate the task as multiple-choice questions for easy and robust evaluation. 
However, our dataset can also be used for generative evaluation, as will be demonstrated in our empirical study. 

To establish baseline performances on \dataname, we explore several state-of-the-art neural models developed for reading comprehension. 
Furthermore, we propose a new  architecture variant that is better suited for commonsense-driven reading comprehension. Still, experimental results demonstrate a significant gap between machine (68.4\% accuracy) and human performance (94.0\%). We provide detailed analysis to provide insights into potentially promising research directions.

\section{Dataset Design}

\subsection{Context Paragraphs} 

We gather a diverse collection of everyday situations from a corpus of personal narratives~\cite{gordon2009identifying} from the Spinn3r Blog Dataset~\cite{burton2009icwsm}. Appendix \ref{appendix:prepro} provides additional details on data pre-processing.

\subsection{Question and Answer Collection}
\label{subsec:qacollect}
We use Amazon Mechanical Turk (AMT) to collect questions and answers. Specifically, for each paragraph, we ask a worker to craft at most two questions that are related to the context and require commonsense knowledge. We encourage the workers to craft questions from but not limited to the following four categories:

\begin{itemize}[leftmargin=9pt,topsep=6pt,itemsep=0pt,parsep=3pt]
\item \textbf{Causes of events:} What may (or may not) be the plausible reason for an event?
\item \textbf{Effects of events:} What may (or may not) happen before (or after, or during) an event?
\item \textbf{Facts about entities:} What may (or may not) be a plausible fact about someone or something?
\item \textbf{Counterfactuals:} What may (or may not) happen if an event happens (or did not happen)?
\end{itemize}

\begin{table*}[!htp]
\small
\centering
\begin{tabular}{l|ccc|c}
\toprule 
 & Train & Dev & Test & All \\ \midrule 
\# Questions (Paragraphs) & 25,588 (13,715) & 3,000 (2,460) & 7,000 (5,711) & 35,588 (21,866) \\
Ave./Max. \# Tokens / Paragraph & 69.4 / 152 & 72.6 / 150 & 73.1 / 149 & 70.3/ 152 \\
Ave./Max. \# Tokens / Question & 10.3 / 34 & 11.2 / 28 & 11.2 / 29 & 10.6 / 34 \\
Ave./Max. \# Tokens / Correct Answer & 8.0 / 40 & 9.7 / 41 & 9.7 / 36 & 8.5 / 41 \\
Ave./Max. \# Tokens / Incorrect Answer & 7.6 / 40 & 9.1 / 38 & 9.1 / 36 & 8.0 / 40 \\ 
Percentage of Unanswerable Questions & 5.9\% & 8.7\% & 8.4\% & 6.7\% \\ 
\toprule
\end{tabular}
\caption{Statistics of training, dev and test sets of \dataname.}
\label{statistics}
\end{table*}

\begin{figure*}[h]
\centering
\subfigure[\dataname]{
\begin{minipage}{0.35\textwidth}
\centering
\includegraphics[width=0.9\textwidth]{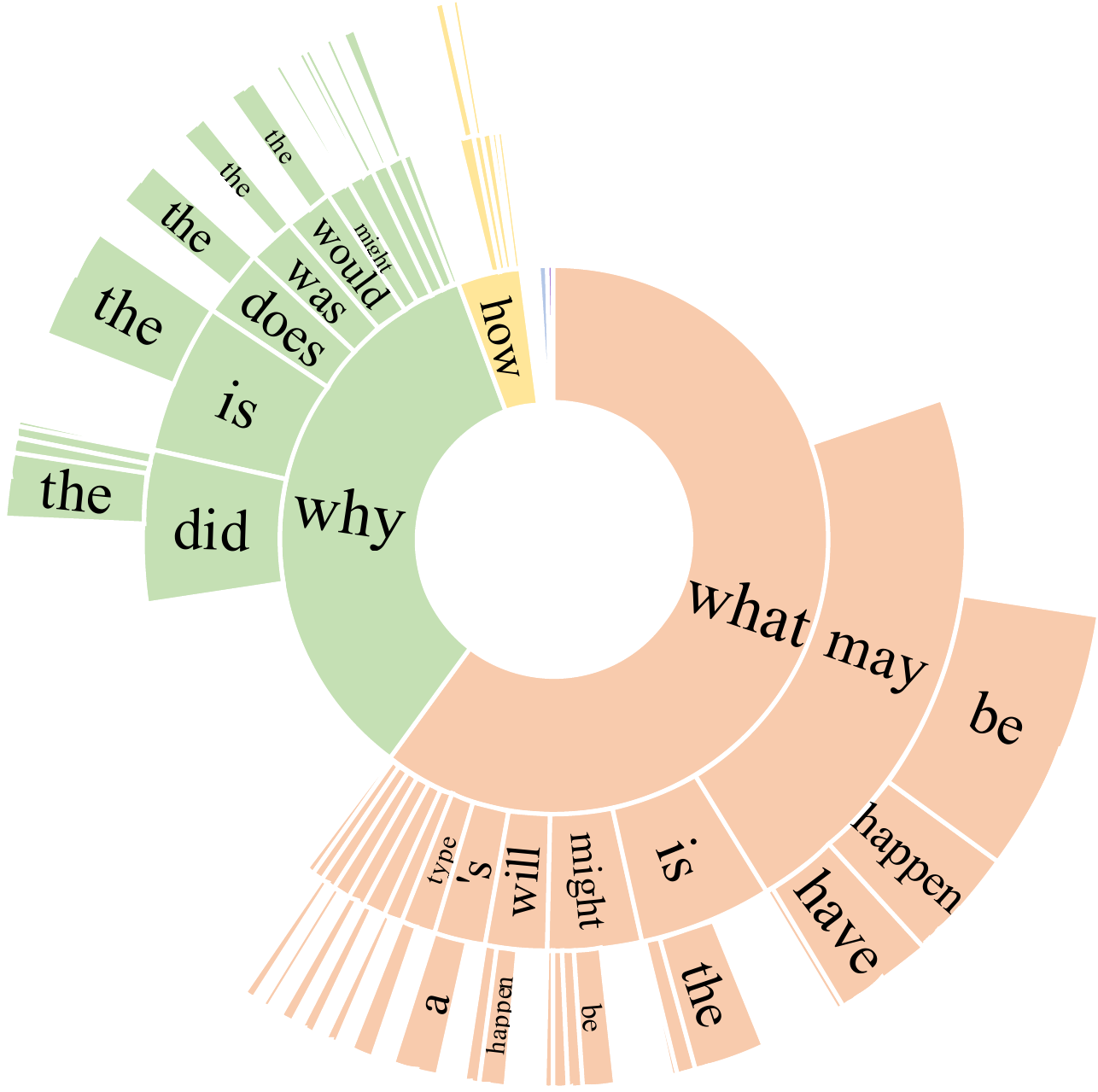}
\end{minipage}
}
\subfigure[SQuAD 2.0]{
\begin{minipage}{0.35\textwidth}
\centering
\includegraphics[width=0.9\textwidth]{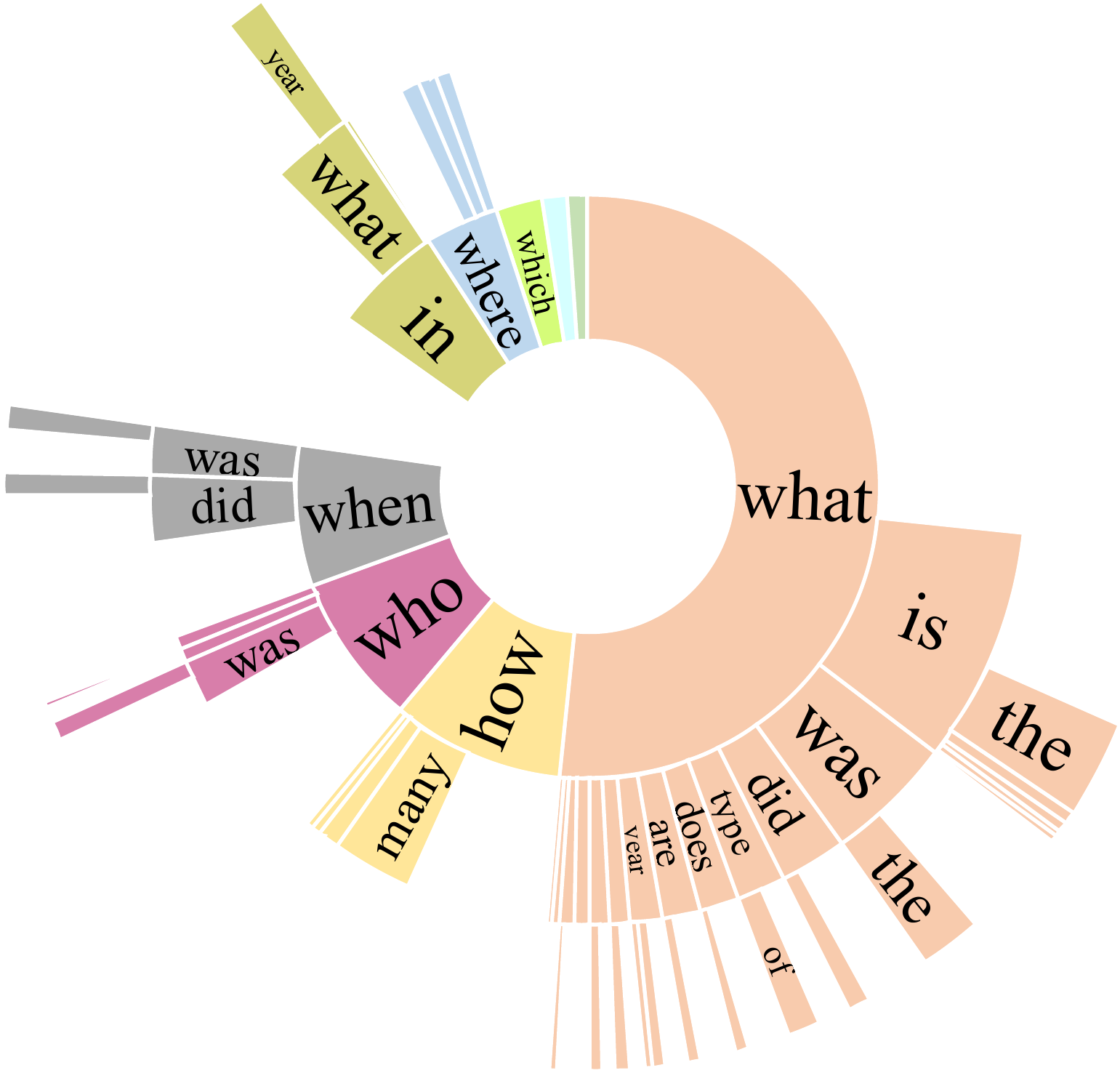}
\end{minipage}
}
\subfigure{
\begin{minipage}{0.07\textwidth}
\centering
\includegraphics[width=0.9\textwidth]{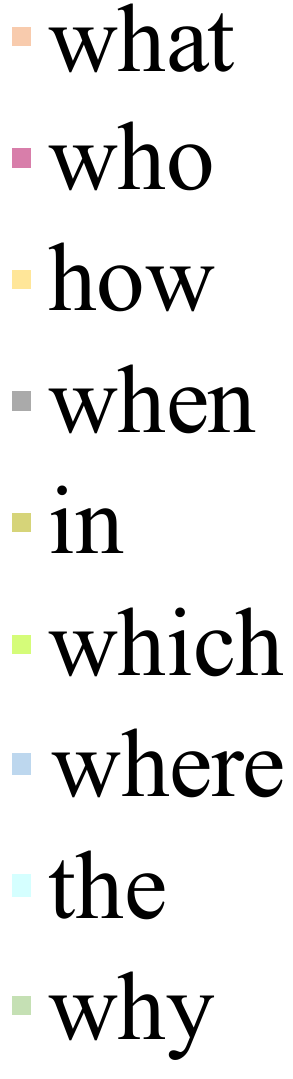}
\end{minipage}
}
\vspace{-0.2cm}
\caption{Distribution of trigram prefixes of questions in \dataname and SQuAD 2.0}
\label{question_dis_figure}
\vspace{-0.3cm}
\end{figure*}

\noindent
These 4 categories of questions literally cover all 9 types of social commonsense of~\newcite{sap2018atomic}. Moreover, the resulting commonsense also aligns with 19 ConceptNet relations, e.g., \textit{Causes}, \textit{HasPrerequisite} and \textit{MotivatedByGoal}, covering about 67.8\% of ConceptNet types. For each question, we also ask a worker to craft at most two correct answers and three incorrect answers. 
We paid workers \$0.7 per paragraph, which is about \$14.8 per hour. Appendix~\ref{appendix:amt_instruction} provides additional details on AMT instructions.

\subsection{Validation}

We create multiple tasks to have humans verify the data. Given a paragraph, a question, a correct answer and three incorrect answers,\footnote{If a question is crafted with two correct answers, we will create two question sets with each correct answer and the same three incorrect answers.} we ask AMT workers to determine the following sequence of questions: 
(1) whether the paragraph is inappropriate or nonsensical, 
(2) whether the question is nonsensical or not related to the paragraph, 
(3) whether they can determine the most plausible correct answer, 
(4) if they can determine the correct answer, whether the answer requires commonsense knowledge, and 
(5) if they can determine the correct answer, whether the answer can be determined without looking at the paragraph. 

We follow the same criterion as in Section~\ref{subsec:qacollect} and ask $3$ workers to work on each question set. Workers are paid $\$0.1$ per question. We consider as valid question set where at least two workers correctly picked the intended answer and all of the workers determined the paragraph/question/answers as satisfactory. Finally we obtain $33,219$ valid question sets in total.

\subsection{Unanswerable Question Creation}

With human validation, we also obtain a set of questions for which workers can easily determine the correct answer without looking at the context or using commonsense knowledge. To take advantage of such questions and encourage AI systems to be more consistent with human understanding, we create unanswerable questions for \dataname. Specifically, from validation outputs, we collect $2,369$ questions for which at least two workers correctly picked the answer and at least on worker determined that it is answerable without looking at the context or requires no common sense. We replace the correct choice of these questions with a ``None of the above'' choice.

To create false negative training instances, we randomly sample $70\%$ of questions from the $33,219$ good question sets and replace their least challenging negative answer with ``None of the above''. Specifically, we fine-tune three BERT\footnote{Through the whole paper, BERT refers to the pre-trained BERT large uncased model from \url{https://github.com/huggingface/pytorch-pretrained-BERT}} next sentence prediction models on \papername: BERT($A|P, Q$), BERT($A|P$), BERT($A|Q$), where $P$, $Q$, $A$ denotes the paragraph, question, and answer. BERT($A|\triangle$) denotes the possibility of an answer $A$ being the next sentence of $\triangle$. The least challenging negative answer is determined by 
\vspace*{-2mm}
\begin{displaymath}
\footnotesize
A^{'} =  \arg\min(\sum_{\forall\triangle\subseteq\{P, Q\}}\text{BERT}(A|\triangle))
\end{displaymath}
\vspace*{-2mm}

\subsection{Train / Dev / Test Split}
We finally obtain $35,588$ question sets for our \papername dataset. To ensure that the development and test sets are of high quality, we identify a group of workers who excelled in the generation task for question and answers, and randomly sample $7$K question sets authored by these excellent workers as test set, and $3$K question sets as development set. The remaining questions are all used as training set. Table~\ref{statistics} shows dataset statistics.

\subsection{Data Analysis}
\label{sec:data_analysis}

\begin{figure*}[h]
\centering
\includegraphics[width=0.99\textwidth]{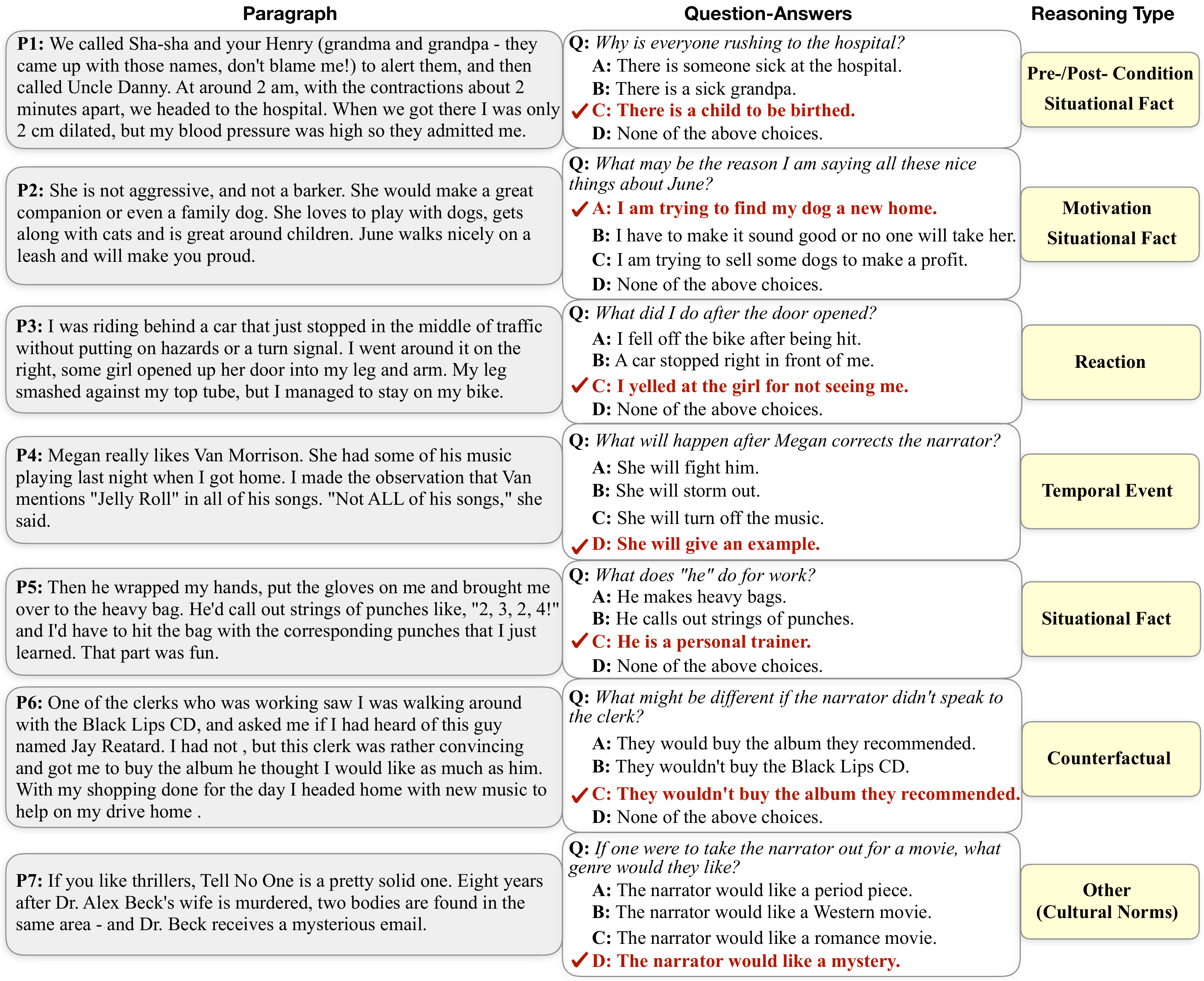}
\caption{Examples of each type of commonsense reasoning in \dataname. ({\cmark} indicates the correct answer.)}
\label{reasoning_category}
\end{figure*}

Figure~\ref{question_dis_figure} compares frequent trigram prefixes in \papername and SQuAD 2.0~\cite{rajpurkar2018know}. Most of the frequent trigram prefixes in \papername, e.g., \textit{why}, \textit{what may happen}, \textit{what will happen} are almost absent from SQuAD 2.0, which demonstrates the unique challenge our dataset contributes. We randomly sample $500$ answerable questions to manually categorize according to their contextual commonsense reasoning types. Figure~\ref{reasoning_category} shows representative examples. Table~\ref{knowledge_type_distribution} shows the distribution of the question types.

\begin{itemize}[leftmargin=9pt,topsep=6pt,itemsep=0pt,parsep=3pt]
\item \textbf{Pre-/post-conditions:} causes/effects of an event.
\item \textbf{Motivations:} intents or purposes. 
\item \textbf{Reactions:} possible reactions of people or objects to an event.
\item \textbf{Temporal events:} what events might happen before or after the current event.
\item \textbf{Situational facts:} facts that can be inferred from the description of a particular situation.
\item \textbf{Counterfactuals:} what might happen given a counterfactual condition.
\item \textbf{Other:} other types, e.g., cultural norms. 
\end{itemize}

\begin{table}[!htp]
\footnotesize
\centering
\begin{tabular}{p{1.0cm}p{3.0cm}|c}
\toprule
\multicolumn{2}{c|}{Type} & Percentage (\%) \\ \midrule
\multicolumn{2}{l|}{MRC w/o commonsense} & 6.2 \\ \midrule
\multicolumn{2}{l|}{MRC w/ commonsense} & 93.8 \\
& Pre-/Post- Condition & \hspace{9mm} 27.2 \\
& Motivation & \hspace{9mm} 16.0 \\  
& Reaction & \hspace{9mm} 13.2 \\
& Temporal Events & \hspace{9mm} 12.4 \\
& Situational Fact & \hspace{9mm} 23.8 \\
& Counterfactual & \hspace{10mm} 4.4 \\
& Other & \hspace{9mm} 12.6 \\
\toprule
\end{tabular}
\vspace*{-3mm}
\caption{The distribution of contextual commonsense reasoning types in \papername.}
\label{knowledge_type_distribution}
\end{table}

\section{Model}
\subsection{BERT with Multiway Attention}

\begin{figure*}[h]
\centering
\includegraphics[width=0.73\textwidth]{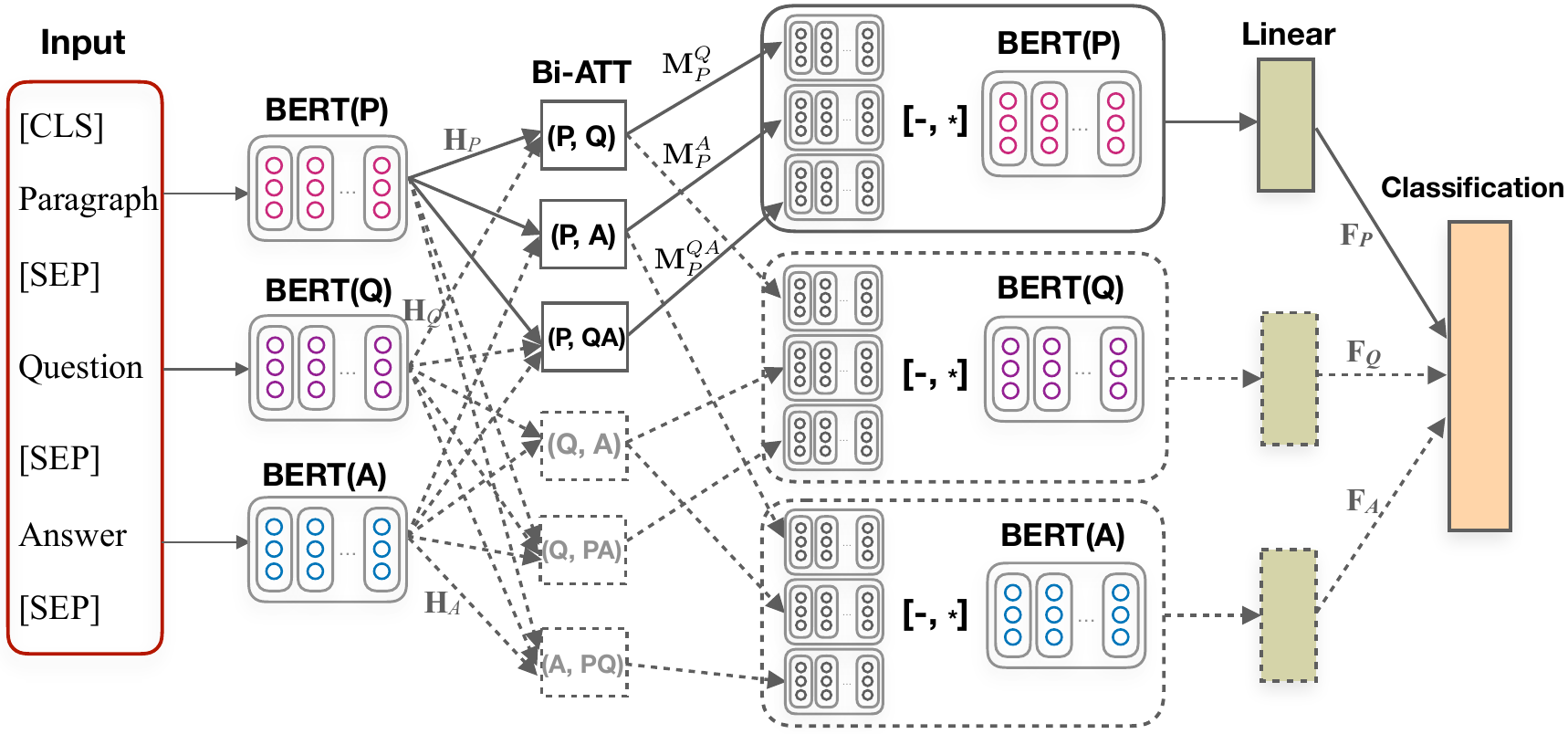}
\caption{Architecture overview of BERT with multiway attention: Solid lines and blocks show the learning of multiway attentive context paragraph representation.}
\label{bert_multiway}
\end{figure*}

Multiway attention~\cite{wang2018yuanfudao,zhu2018hierarchical} has been shown to be effective in capturing the interactions between each pair of input paragraph, question and candidate answers, leading to better context interpretation, while BERT fine-tuning~\cite{devlin2018bert} also shows its prominent ability in commonsense inference. To further enhance the context understanding ability of BERT fine-tuning, we perform multiway bidirectional attention over the BERT encoding output. Figure~\ref{bert_multiway} shows the overview of the architecture. 

\paragraph{Encoding with Pre-trained BERT} 
Given a paragraph, a question, and a set of candidate answers, the goal is to select the most plausible correct answer from the candidates. We formulate the input paragraph as $P = \{p_0, p_1, ..., p_n\}$, the question as $Q=\{q_0, q_1, ..., q_k\}$ and a candidate answer as $A=\{a_0, a_1, ..., a_s\}$, where $p_i$, $q_i$ and $a_i$ is the $i$-th word of the paragraph, question and candidate answer respectively. Following~\cite{devlin2018bert}, given the input $P$, $Q$ and $A$, we apply the same tokenizer and concatenate all tokens as a new sequence $\big[\text{[CLS]}, P, \text{[SEP]}, Q, \text{[SEP]}, A, \text{[SEP]}\big]$, where [CLS] is a special token used for classification and [SEP] is a delimiter. Each token is initialized with a vector by summing the corresponding token, segment and position embedding from pre-trained BERT, and then encoded into a hidden state. 
Finally we get $[\boldsymbol{\mathrm{H}}_{cls}, \boldsymbol{\mathrm{H}}_{P}, \boldsymbol{\mathrm{H}}_{Q}, \boldsymbol{\mathrm{H}}_{A}]$ as encoding output.

\paragraph{Multiway Attention}
To encourage better context interpretation, we perform multiway attention over BERT encoding output. Taking the paragraph $P$ as an example, we compute three types of attention weights to capture its correlation to the question, the answer, and both the question and answer, and get question-attentive, answer-attentive, and question and answer-attentive paragraph representations
\begin{displaymath}
\small
\tilde{\boldsymbol{\mathrm{H}}}_{P} = \boldsymbol{\mathrm{H}}_{P}\boldsymbol{W}_t + \boldsymbol{b}_t
\end{displaymath}
\begin{displaymath}
\small
\boldsymbol{\mathrm{M}}_{P}^{Q} = \mathrm{Softmax}(\tilde{\boldsymbol{\mathrm{H}}}_{P}\boldsymbol{\mathrm{H}}_{Q}^\top)\boldsymbol{\mathrm{H}}_{Q}
\end{displaymath}
\begin{displaymath}
\small
\boldsymbol{\mathrm{M}}_{P}^{A} = \mathrm{Softmax}(\tilde{\boldsymbol{\mathrm{H}}}_{P}\boldsymbol{\mathrm{H}}_{A}^\top)\boldsymbol{\mathrm{H}}_{A}
\end{displaymath}
\begin{displaymath}
\small
\boldsymbol{\mathrm{M}}_{P}^{QA} = \mathrm{Softmax}(\tilde{\boldsymbol{\mathrm{H}}}_{P}\boldsymbol{\mathrm{H}}_{QA}^\top)\boldsymbol{\mathrm{H}}_{QA}
\end{displaymath}
where $\boldsymbol{W}_t$ and $\boldsymbol{b}_t$ are learnable parameters. Next we fuse these representations with the original encoding output of $P$
\begin{displaymath}
\small
\boldsymbol{\mathrm{F}}_{P}^{Q} = \sigma([\boldsymbol{\mathrm{H}}_{P}\boldsymbol{\mathrm{M}}_{P}^{Q}: \boldsymbol{\mathrm{H}}_{P}-\boldsymbol{\mathrm{M}}_{P}^{Q}]\boldsymbol{W}_{P} + \boldsymbol{b}_{P})
\end{displaymath}
\begin{displaymath}
\small
\boldsymbol{\mathrm{F}}_{P}^{A} = \sigma([\boldsymbol{\mathrm{H}}_{P}\boldsymbol{\mathrm{M}}_{P}^{A}: \boldsymbol{\mathrm{H}}_{P}-\boldsymbol{\mathrm{M}}_{P}^{A}]\boldsymbol{W}_{P} + \boldsymbol{b}_{P})
\end{displaymath}
\begin{displaymath}
\small
\boldsymbol{\mathrm{F}}_{P}^{QA} = \sigma([\boldsymbol{\mathrm{H}}_{P}\boldsymbol{\mathrm{M}}_{P}^{QA}: \boldsymbol{\mathrm{H}}_{P}-\boldsymbol{\mathrm{M}}_{P}^{QA}]\boldsymbol{W}_{P} + \boldsymbol{b}_{P})
\end{displaymath}
where $[:]$ denotes concatenation operation. $\boldsymbol{W}_P$, $\boldsymbol{b}_P$ are learnable parameters for fusing paragraph representations. $\sigma$ denotes ReLU function.

Finally, we apply column-wise max pooling on $[\boldsymbol{\mathrm{F}}_{P}^{Q}: \boldsymbol{\mathrm{F}}_{P}^{A}: \boldsymbol{\mathrm{F}}_{P}^{QA}]$ and obtain the new paragraph representation $\boldsymbol{\mathrm{F}}_{P}$. Similarly, we can also obtain a new representation $\boldsymbol{\mathrm{F}}_Q$ and $\boldsymbol{\mathrm{F}}_A$ for $Q$ and $A$ respectively. We use $\boldsymbol{\mathrm{F}} = [\boldsymbol{\mathrm{F}}_P: \boldsymbol{\mathrm{F}}_Q: \boldsymbol{\mathrm{F}}_A]$ as the overall vector representation for the set of paragraph, question and a particular candidate answer.

\paragraph{Classification}
For each candidate answer $A_i$, we compute the loss as follows:
\begin{displaymath}
\small
L(A_i|P, Q) = -\log\frac{\exp(\boldsymbol{W}_{f}^{\top}\boldsymbol{\mathrm{F}}_i)}{\sum_{j=1}^{4}\exp(\boldsymbol{W}_{f}^{\top}\boldsymbol{\mathrm{F}}_j))}
\end{displaymath}

\section{Experiments}
\subsection{Baseline Methods}
We explore two categories of baseline methods: reading comprehension approaches and pre-trained language model based approaches.

\begin{table*}[!htp]
\footnotesize
\centering
\begin{tabular}{l|cccc|>{\columncolor[gray]{0.95}}c>{\columncolor[gray]{0.95}}c}
\toprule 
Model & Att(P, Q) & Att(P, A) & Att(Q, A) & Pre-training LM &  Dev & Test \\ \midrule
Sliding Window~\cite{richardson2013mctest} & \xmark & \xmark & \xmark & \xmark & 25.0 & 24.9\\
Stanford Attentive Reader~\cite{chen2016thorough} & UD & \xmark & \xmark & \xmark & 45.3 & 44.4 \\
Gated-Attention Reader~\cite{dhingra2017gated} & Multi-hop UD & \xmark & \xmark & \xmark & 46.9 & 46.2 \\
Co-Matching~\cite{wang2018co} & UD & UD & \xmark & \xmark & 45.9 & 44.7 \\
Commonsense-Rc~\cite{wang2018yuanfudao} & UD & UD & UD & \xmark & 47.6 & 48.2 \\  
GPT-FT~\cite{radford2018improving} & \xmark & \xmark & \xmark & UD & 54.0 & 54.4 \\
BERT-FT~\cite{devlin2018bert} & \xmark & \xmark & \xmark & BD & 66.2 & 67.1 \\ 
DMCN~\cite{zhang2019dual} & UD & UD & \xmark & BD & 67.1 & 67.6 \\ \midrule 
BERT-FT Multiway & BD & BD & BD & BD & 68.3 & 68.4 \\ \midrule
Human & & & & & & 94.0 \\ \toprule 
\end{tabular}
\vspace{-0.1cm}
\caption{Comparison of varying approachs (Accuracy \%). Att: Attention, UD: Unidirectional, BD: Bidirectional}
\label{results}
\vspace{-0.2cm}
\end{table*}

\begin{itemize}[leftmargin=0pt,topsep=3pt,itemsep=0pt,parsep=3pt]
\item[] \textbf{Sliding Window}~\cite{richardson2013mctest} measures the similarity of each candidate answer with each window with $m$ words of the paragraph.
\item[] \textbf{Stanford Attentive Reader}~\cite{chen2016thorough} performs a bilinear attention between the question and paragraph for answer prediction.
\item[] \textbf{Gated-Attention Reader}~\cite{dhingra2017gated} performs multi-hop attention between the question and a recurrent neural network based paragraph encoding states.
\item[] \textbf{Co-Matching}~\cite{wang2018co} captures the interactions between question and paragraph, as well as answer and paragraph with attention.
\item[] \textbf{Commonsense-RC}~\cite{wang2018yuanfudao} applies three-way unidirectional attention to model interactions between paragraph, question, and answers.
\item[] \textbf{GPT-FT}~\cite{radford2018improving} is based on a generative pre-trained transformer language model, following a fine-tuning step on \dataname.
\item[] \textbf{BERT-FT}~\cite{devlin2018bert} is a pre-trained bidirectional transformer language model following a fine-tuning step on \dataname.
\item[] \textbf{DMCN}~\cite{zhang2019dual} performs dual attention between paragraph and question/answer over BERT encoding output.
\end{itemize}

\paragraph{Human Performance} To get human performance on \papername QA, we randomly sample 200 question sets from the test set, and ask $3$ workers from AMT to select the most plausible correct answer. Each worker is paid $\$0.1$ per question set. We finally combine the predictions for each question with majority vote. 

\subsection{Results and Analysis}

Table~\ref{results} shows the characteristics and performance of varying approaches and human performance.\footnote{Appendix~\ref{appendix:implementation} shows the implementation details.}

\begin{figure}[!htp]
\centering
\includegraphics[width=0.5\textwidth]{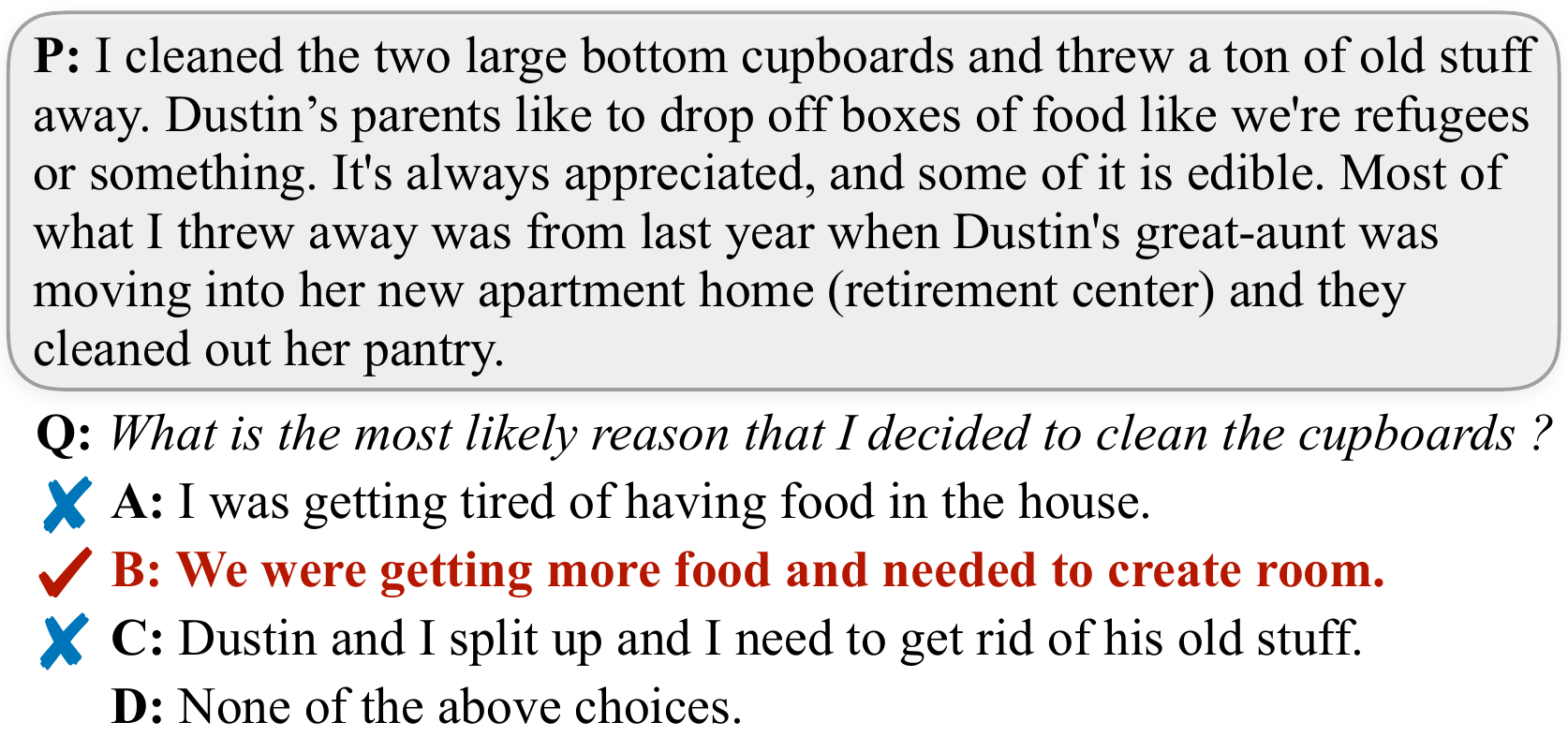}
\caption{Prediction comparison between our approach (B) with Commonsense-RC (C) and BERT-FT (A).}
\label{fig:error_comparison}
\vspace{-0.2cm}
\end{figure}

Most of the reading comprehension approaches apply attention to capture the correlation between paragraph, question and each candidate answer and tend to select the answer which is the most semantically closed to the paragraph. For example, in Figure~\ref{fig:error_comparison}, the Commonsense-RC baseline mistakenly selected the choice which has the most overlapped words with the paragraph without any commonsense reasoning. However, our analysis shows that more than 83\% of correct answers in \dataname are not stated in the given paragraphs, thus simply comparing the semantic relatedness doesn't work well. Pre-trained language models with fine-tuning achieve more than 20\% improvement over reading comprehension approaches. 

By performing attention over BERT-FT, the performance is further improved, which demonstrates our assumption that incorporating interactive attentions can further enhance the context interpretation of BERT-FT. For example, in Figure~\ref{fig:error_comparison}, BERT-FT mistakenly selected choice A which can be possibly entailed by the paragraph. However, by performing multiway attention to further enhance the interactive comprehension of context, question and answer, our approach successfully selected the correct answer.

\section{Discussion}

\subsection{Ablation Study} 

\begin{table}[!htp]
\small
\centering
\begin{tabular}{l|cc}
\toprule 
Model & Dev Acc (\%) & Test Acc (\%)  \\ \midrule 
BERT-FT (A$|$P, Q) & 66.2 & 67.1 \\ 
BERT-FT (A$|$P) & 63.5 & 64.5 \\
BERT-FT (A$|$Q) & 56.2 & 55.9 \\ 
BERT-FT (A) & 40.3 & 40.3 \\ \toprule 
\end{tabular}
\caption{Ablation of Paragraphs (P) or Questions (Q)}
\label{ablation_results}
\end{table}

Many recent studies have suggested the importance of measuring the dataset bias by checking the model performance based on partial information of the problem~\cite{gururangan2018annotation,cai2017pay}. Therefore, we report problem ablation study in Table~\ref{ablation_results} using BERT-FT as a simple but powerful straw man approach.  
Most notably, ablating questions does not cause significant performance drop. Further investigation indicates that this is because the high-level question types, e.g., \textit{what happens next}, \textit{what happened before}, are not diverse, so that
it is often possible to make a reasonable guess on what the question may have been based on the context and the answer set. Ablating other components of the problems cause more significant drops in performance.

\subsection{Knowledge Transfer Through Fine-tuning}
Recent studies~\cite{howard2018universal,min2017question,devlin2018bert} have shown the benefit of fine-tuning on similar tasks or datasets for knowledge transfer. Considering the unique challenge of \papername, we explore two related multiple-choice datasets for knowledge transfer: RACE~\cite{lai2017race}, a large-scale reading comprehension dataset, and SWAG~\cite{zellers2018swag}, a large-scale commonsense inference dataset. Specifically, we first fine-tune BERT on RACE or SWAG or both, and directly test on \papername to show the impact of knowledge transfer. Furthermore, we sequentially fine-tune BERT on both RACE or SWAG and \papername. As Table~\ref{related-ft-results} shows, with direct knowledge transfer, RACE provides significant benefit than SWAG since \papername requires more understanding of the interaction between paragraph, question and each candidate answer. With sequentially fine-tuning, SWAG provides better performance, which indicates that with fine-tuning on SWAG, BERT can obtain better commonsense inference ability, which is also beneficial to \papername. 

\begin{table}[!htp]
\small
\centering
\begin{tabular}{l|cc}
\toprule 
Model & Dev Acc  & Test Acc \\ \midrule 
BERT-FT$_{\text{SWAG}}$ & 28.9 & 28.5 \\
BERT-FT$_{\text{RACE}}$ & 42.0 & 42.5 \\
BERT-FT$_{\text{RACE+SWAG}}$ & 44.2 & 45.1 \\ \midrule
BERT-FT$_{\text{SWAG}\to\text{Cosmos}}$ & 67.8 & 68.9 \\ 
BERT-FT$_{\text{RACE}\to\text{Cosmos}}$ & 67.4 & 68.2 \\
BERT-FT$_{\text{RACE+SWAG}\to\text{Cosmos}}$ & 67.1 & 68.7 \\ \toprule 
\end{tabular}
\caption{Knowledge transfer through fine-tuning. (\%)}
\label{related-ft-results}
\end{table}

\subsection{Error Analysis}
We randomly select 100 errors made by our approach from the dev set, and identify 4 phenomena:

\begin{figure}[t]
\centering
\includegraphics[width=0.5\textwidth]{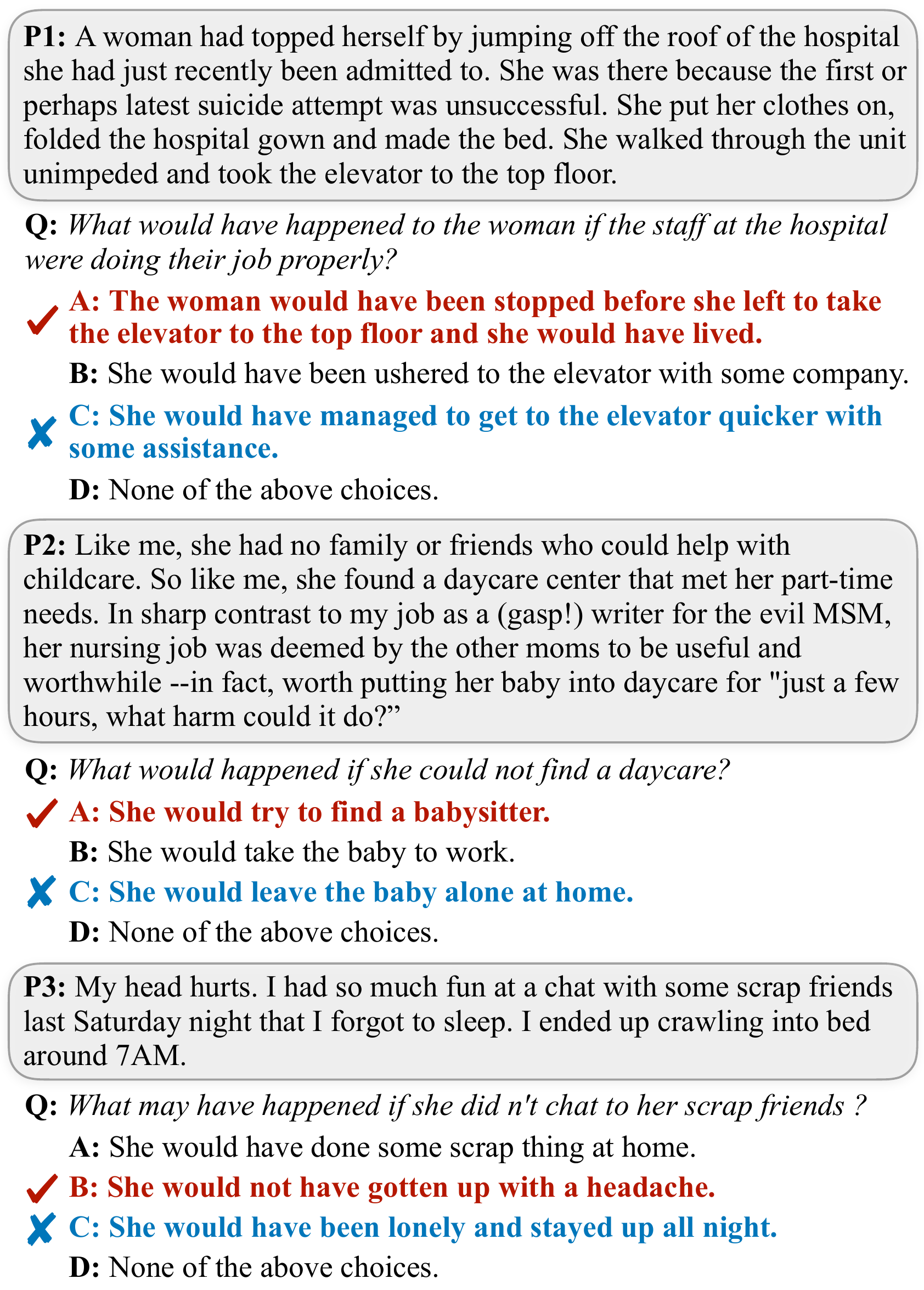}
\caption{Examples errors of our approach. ({\cmark} indicates correct answers and {\xmark} shows prediction errors.)}
\label{fig:error}
\vspace{-0.6cm}
\end{figure}

\begin{itemize}[leftmargin=0pt,topsep=3pt,itemsep=0pt,parsep=3pt]
\item[] \textbf{Complex Context Understanding:} In 30\% of the errors, the context requires complicated cross-sentence interpretation and reasoning. Taking P1 in Figure~\ref{fig:error} as an example, to correctly predict the answer, we need to combine the context information that \textit{the woman attempted to suicide before but failed}, \textit{she made the bed since she determined to leave}, and \textit{she took the elevator and headed to the roof}, and infer that \textit{the woman was attempting to suicide again}.
\item[] \textbf{Inconsistent with Human Common Sense:} In 33\% of the errors, the model mistakenly selected the choice which is not consistent with human common sense. For example, in P2 of Figure~\ref{fig:error}, both choice A and choice C could be potentially correct answers. However, from human common sense, \textit{it's not safe to leave a baby alone at home}.
\item[] \textbf{Multi-turn Commonsense Inference:} 19\% of the errors are due to multi-turn commonsense inference. For example, in P3 of Figure~\ref{fig:error}, the model needs to first determine the cause of \textit{headache} is that \textit{she chatted with friends and forgot to sleep} using common sense. Further, with counterfactual reasoning, if \textit{she didn't chat to her friends},  then \textit{she wouldn't have gotten up with a headache}.
\item[] \textbf{Unanswerable Questions:} 14\% of the errors are from unanswerable questions. The model cannot handle ``None of the above'' properly since it cannot be directly entailed by the given paragraph or the question. Instead, the model needs to compare the potential of all the other candidate choices.
\end{itemize}

\subsection{Generative Evaluation}
\begin{figure}[!htp]
\centering
\includegraphics[width=0.49\textwidth]{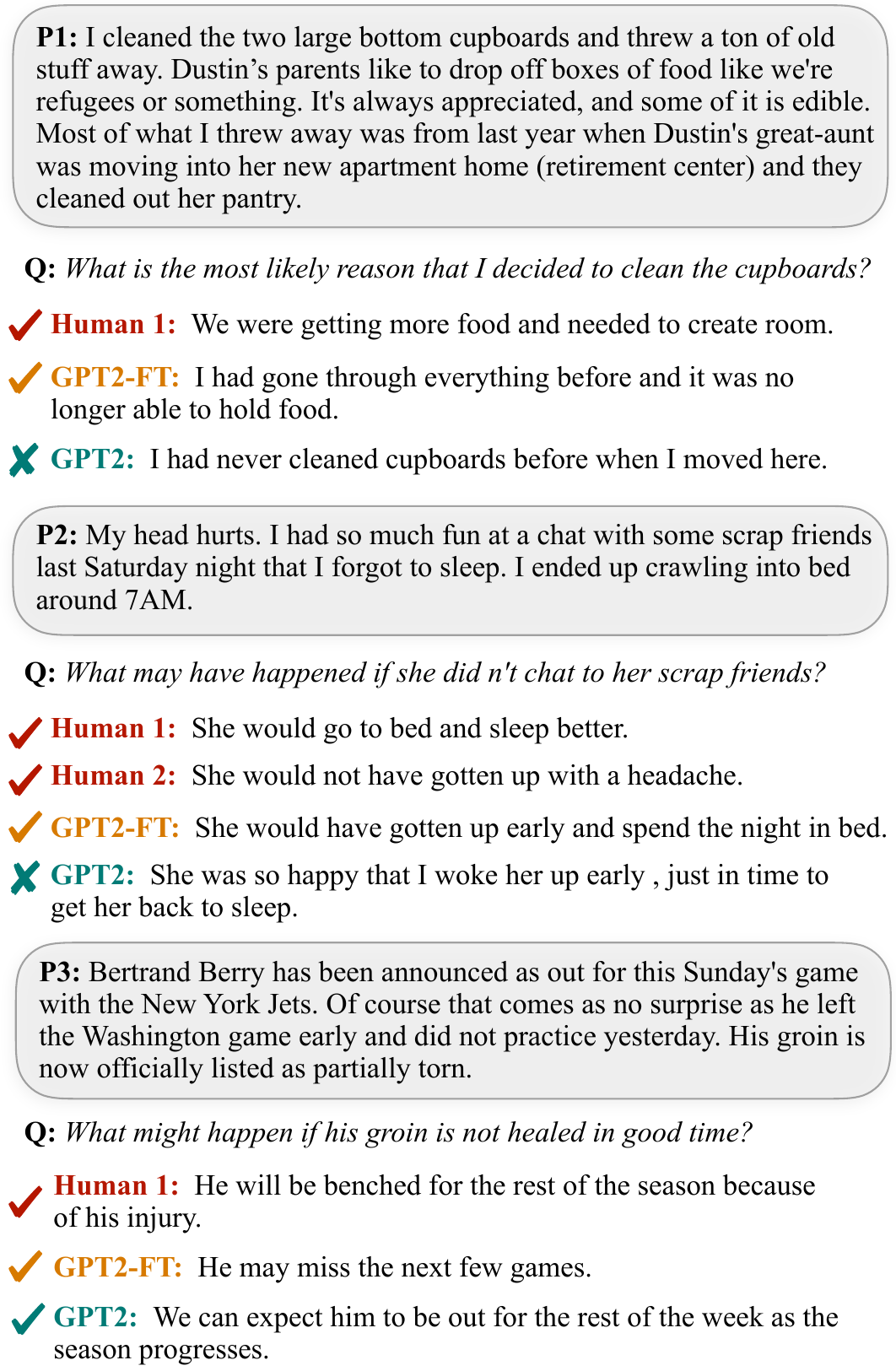}
\caption{Examples of human authored correct answers, and automatically generated answers by pre-trained GPT2 and GPT2-FT. (\cmark indicates the answer is correct while \xmark shows that the answer is incorrect.)}
\label{generative_examples}
\vspace{-0.2cm}
\end{figure}

In real world, humans are usually asked to perform contextual commonsense reasoning without being provided with any candidate answers. To test machine for human-level intelligence, we leverage a state-of-the-art natural language generator GPT2~\cite{radford2019language} to automatically generate an answer by reading the given paragraph and question. Specifically, we fine-tune a pre-trained GPT2 language model on all the $\big[\textit{Paragraph}, \textit{Question}, \textit{Correct Answer}\big]$ of \papername training set, then given each $\big[\textit{Paragraph}, \textit{Question}\big]$ from test set, we use GPT2-FT to generate a plausible answer. We automatically evaluate the generated answers against human authored correct answers with varying metrics in Table~\ref{generative-eval}. We also create a AMT task to have $3$ workers select all plausible answers among $4$ automatically generated answers and a ``None of the aboce'' choice for 200 question sets. We consider an answer as correct only if all 3 workers determined it as correct. Figure~\ref{generative_examples} shows examples of automatically generated answers by pre-trained GPT2 and GPT2-FT as well as human authored correct answers. We observe that by fine-tuning on \papername, GPT2-FT generates more accurate answers. Although intuitively there may be multiple correct answers to the questions in \dataname, our analysis shows that more than 84\% of generated correct answers identified by human are semantically consistent with the gold answers in \papername, which demonstrates that \papername can also be used as a benchmark for generative commonsense reasoning. Appendix~\ref{appendix:generative} shows more details and examples for generative evaluation.

\begin{table}[!htp]
\footnotesize
\centering
\begin{tabular}{p{4.8cm}p{0.5cm}<{\centering}p{1.3cm}<{\centering}}
\toprule 
Metrics & GPT2 & GPT2-FT \\ \midrule
BLEU~\cite{papineni2002bleu} & 10.7  &  21.0 \\
METEOR~\cite{banerjee2005meteor} & 7.2 &  8.6 \\
ROUGE-L~\cite{lin2004rouge}  & 13.9 & 22.1 \\
CIDEr~\cite{vedantam2015cider} & 0.05  & 0.17 \\
BERTScore F1~\cite{zhang2019bertscore} & 41.9 & 44.5 \\ \midrule
Human & 11.0\% & 29.0\% \\ \toprule
\end{tabular}
\caption{Generative performance of pre-trained GPT2 and GPT2-FT on \dataname. All automatic metric scores are averaged from 10 sets of sample output.}
\label{generative-eval}
\vspace{-0.2cm}
\end{table}

\begin{table*}[!htp]
\footnotesize
\centering
\begin{tabular}{p{2.0cm}p{0.7cm}<{\centering}p{0.7cm}<{\centering}p{1.6cm}<{\centering}p{2.8cm}<{\centering}p{2.0cm}<{\centering}p{0.8cm}<{\centering}p{2cm}<{\centering}p{1.5cm}<{\centering}}
\toprule
Dataset & Size & Type & Answer Type & Paragraph Source & Questions/ Answers & Require MRC & Require Common Sense \\ \midrule 
MCTest & 2K & PQA  & MC & MTurk & MTurk & \cmark & - \\
RACE & 100K & PQA & MC & Human Experts & Human Experts & \cmark & - \\ 
MCScript & 13.9K & PQA & MC & MTurk & MTurk  & \cmark & 27.4\% \\ 
NarrativeQA & 46.8K & PQA & Open Text & Books/Movie Scripts & MTurk & \cmark & - \\
ARC & 7.8K & QA & MC & N/A & Web & \xmark & - \\
CommonsenseQA & 12.2K & QA & MC & N/A & MTurk/Web & \xmark & 100\% \\
ReCoRD & 121K & PQA & Span & News & Automatic  & \cmark & 75.0\% \\
\papername & 31.8K & PQA & MC & Webblog & MTurk & \cmark & 93.8\% \\ \toprule 
\end{tabular}
\caption{Comparison of the \papername QA to other multiple-choice machine reading comprehension datasets: P: contextual paragraph, Q: question, A: answers, MC: Multiple-choice, and - means unknown.}
\label{dataset:comparison}
\end{table*}

\section{Related Work}

There have been many exciting new datasets developed for reading comprehension,  
such as 
SQuAD~\cite{rajpurkar2016squad}, NEWSQA~\cite{trischler2017newsqa}, SearchQA~\cite{dunn2017searchqa}, NarrativeQA~\cite{kovcisky2018narrativeqa}, ProPara~\cite{mishra2018tracking}, CoQA~\cite{reddy2018coqa},  ReCoRD~\cite{zhang2018record}, Dream~\cite{sun2019dream},  MCTest~\cite{richardson2013mctest}, RACE~\cite{lai2017race},
CNN/Daily Mail~\cite{hermann2015teaching},
Children's Book Test~\cite{hill2015goldilocks}, and MCScript~\cite{ostermann2018mcscript}. 
Most these datasets focus on relatively explicit understanding of the context paragraph, thus a relatively small or unknown fraction of the dataset requires commonsense reasoning, if at all. 

A notable exception is ReCoRD~\cite{zhang2018record} that is designed specifically for challenging reading comprehension with commonsense reasoning.
\papername complements ReCoRD with  three unique challenges: (1) our context is from webblogs rather than news, thus requiring commonsense reasoning for \emph{everyday events} rather than \emph{news-worthy events}. (2) All the answers of ReCoRD are contained in the paragraphs and are assumed to be entities. In contrast, in \papername, more than 83\% of answers are \emph{not} stated in the paragraphs, creating unique modeling challenges. (3) \papername can be used for \emph{generative} evaluation in addition to multiple-choice evaluation.

There also have been other datasets focusing specifically on question answering with commonsense, such as CommonsenseQA~\cite{talmor2018commonsenseqa} and Social IQa~\cite{sap2019socialiqa}, and various other types of commonsense inferences  \cite{levesque2012winograd,rahman2012resolving,gordon2016commonsense,rashkin2018modeling,roemmele2011choice,mostafazadeh2017lsdsem,zellers2018swag}. 
The unique contribution of 
\papername is combining reading comprehension with commonsense reasoning, requiring \emph{contextual} commonsense reasoning over considerably more complex, diverse, and longer context.  Table~\ref{dataset:comparison} shows comprehensive comparison among the most relevant datasets.

There have been a wide range of attention mechanisms developed for reading comprehension datasets
~\cite{hermann2015teaching,kadlec2016text,chen2016thorough,dhingra2017gated,seo2016bidirectional,wang2018co}. Our work investigates various state-of-the-art approaches to reading comprehension, and provide empirical insights into the  design choices that are the most effective for contextual commonsense reasoning required for \papername. 

\section{Conclusion}
We introduced \dataname, a large-scale dataset for machine comprehension with contextual commonsense reasoning. We also presented extensive empirical results comparing various state-of-the-art neural architectures to reading comprehension, and demonstrated a new model variant that leads to the best result. The substantial headroom (25.6\%) between the best model performance and human encourages future research on contextual commonsense reasoning. 

\section*{Acknowledgments}
We thank Scott Yih, Dian Yu, Wenpeng Yin, Rowan Zellers, and anonymous reviewers for helpful discussions and comments. This research was supported in part by NSF (IIS-1524371, IIS-1714566), DARPA under the CwC program through the ARO (W911NF-15-1-0543), DARPA under the MCS program through NIWC Pacific (N66001-19-2-4031), and Allen Institute for AI.

\bibliography{emnlp-ijcnlp-2019}
\bibliographystyle{acl_natbib}

\appendix
\newpage

\section{Context Paragraph Extraction}\label{appendix:prepro}
We gather a diverse collection of everyday situations from a corpus of personal narratives~\cite{gordon2009identifying} from the ICWSM 2009 Spinn3r Blog Dataset~\cite{burton2009icwsm}. 
It contains over $1.6$ million of non-spam weblog entries describing everyday personal events. For each personal story, we use spaCy\footnote{\url{https://spacy.io/}} for sentence segmentation and tokenization. In order to get a short sub-story as context, we apply pre-trained BERT\footnote{Through the whole paper, BERT refers to the pre-trained BERT large uncased model from \url{https://github.com/huggingface/pytorch-pretrained-BERT}}~\cite{devlin2018bert} model to predict a confidence score for each two consecutive sentences from the story, and then segment each story into multiple paragraphs so that each paragraph contains between 30 and 150 words. For each blog, we randomly sample one paragraph as context to create questions and answers.

\section{Additional Details on AMT Instructions}\label{appendix:amt_instruction}
To make the questions more challenging for an AI system, we recommend the workers use less words from the paragraph for correct answers and make incorrect answers more appealing by using words from the paragraph as much as possible. We also encourage workers to provide all the candidate answers with similar length and style.

We restrict this task to the workers in English-speaking countries (United States, Canada, and United Kingdom) and with more than 5,000 HITs with at least a 99\% acceptance rate. To ensure quality, we also create a qualification task. 

\section{Implementation Details}\label{appendix:implementation}
For baseline methods, we use their released implementations from open source projects and retrain them on our dataset. All approaches follow the same pre-processing steps: segment each paragraph into multiple sentences, and tokenize each sentence as well as question and candidate answers with spaCy. For BERT-FT based approaches, we optimize the parameters with grid search: training epochs 10, learning rate $l$ $\in$ \{2e-5, 3e-5, 5e-5\}, gradient accumulation steps $g$ $\in$ \{1, 4, 8\}, training batch size $b$ $\in$ \{2$g$, 3$g$, 4$g$, 5$g$\}. We will make all the resources and implementations publicly available.

\section{Impact of Training Data Size}\label{appendix:learning_curve}
\begin{figure}[!htp]
\centering
\includegraphics[width=0.45\textwidth]{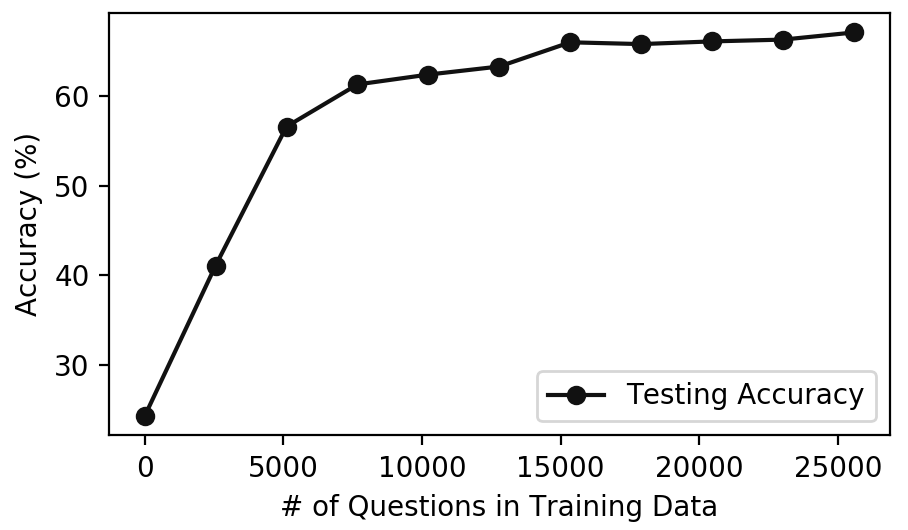}
\caption{Performance on \papername with various amount of training data}
\label{learning_curve}
\end{figure}

To explore the impact of the amount of training data, we divide the whole training dataset into 10-fold and successively add another 10\% into the training data. We use BERT-FT approach for comparison. Figure~\ref{learning_curve} shows the learning curve. We can see that, the performance goes up as we add more training data. However, we do not observe significant improvement when we further increase the training data after $15$K questions.

\section{Details for Generative Evaluation}\label{appendix:generative}

For generative evaluation, we base on the OpenAI pre-trained GPT2 transformer language model,\footnote{\url{https://github.com/huggingface/pytorch-pretrained-BERT}} which has $117$M parameters, and fine-tune it with all \big[\textit{Paragraph}, \textit{Question}, \textit{Correct Answer}\big] in \dataname training set with top-$k$ sampling, where $k \in \{3, 10, 50, 100, 1000\}$.  After fine-tuning, we use GPT2-FT to generate 10 candidate answers conditioned on each \big[\textit{Paragraph}, \textit{Question}\big] from development and test sets. Note that for all training, development and test sets, we omit the questions to which the correct answer in \dataname is ``None of the above''.

For automatic evaluation, we compare each generated candidate answer against the original human authored correct choice in \dataname, and average all metric scores with $10$ sets of candidate answers. For AMT based human evaluation, we randomly sample $200$ paragraphs and questions, and for each question we randomly sample $4$ automatically generated answers from the outputs of GPT2 without fine-tuning and GPT2-FT. For each question set, we ask $3$ workers to select all plausible correct answers from the $4$ candidate choices or ``None of the above''. Workers are paid $\$0.1$ per question set. For each question, we consider an automatically generated answer as correct only if all $3$ workers determined it as correct.

\end{document}